\documentclass[letterpaper]{article} 
\usepackage{amsmath,amssymb,amsfonts,bm} 
\usepackage{amsthm}
\usepackage{aaai25}  
\usepackage{times}  
\usepackage{helvet}  
\usepackage{courier}  
\usepackage[hyphens]{url}  
\usepackage{graphicx} 
\usepackage{natbib}  
\usepackage{caption} 
\usepackage{algorithm}
\usepackage{algorithmic}

\usepackage[capitalize]{cleveref}
\crefname{section}{Sec.}{Secs.}
\Crefname{section}{Section}{Sections}
\Crefname{table}{Table}{Tables}
\crefname{table}{Tab.}{Tabs.}

\usepackage{soul}

\usepackage{booktabs}
\usepackage[switch]{lineno}
\usepackage[table]{xcolor}
\usepackage{makecell}
\graphicspath{{Figures/}}
\usepackage{threeparttable}
\usepackage[accsupp]{axessibility}
\usepackage[utf8]{inputenc}
\usepackage{microtype}
\usepackage{inconsolata}
\usepackage{color}
\usepackage{pgfplots}
\pgfplotsset{compat=1.18} 
\usepackage{latexsym}
\usepackage[caption=false,font=normalsize,
   labelfont=sf,textfont=sf]{subfig}
\usepackage{textcomp}
\usepackage{verbatim}
\usepackage{paralist}
\usepackage{colortbl}
\usepackage{arydshln}
\usepackage{multirow}
\usepackage{tabularx}
\usepackage{pifont}
\usepackage{tcolorbox}
\usepackage[framemethod=tikz]{mdframed}
\usepackage{tikz}
\usepgfplotslibrary{groupplots}
\usepackage{pgf-pie}
\usepackage{filecontents}
\usetikzlibrary{patterns,shapes.geometric,calc}
\usepackage{subcaption}
\usepackage{radar}
\usepackage{array}
\usepackage{fontawesome}
\usepackage{bibentry}

\newcommand{\uline}[1]{
   \tikz[baseline=(todotted.base)]{
        \node[inner sep=0pt, outer sep=0pt] (todotted) {#1};
        \draw[black, thick] (todotted.south west) -- (todotted.south east);
    }%
}

\usepackage{newfloat}
\usepackage{listings}
\DeclareCaptionStyle{ruled}{labelfont=normalfont,labelsep=colon,strut=off} 
\floatstyle{ruled}
\newfloat{listing}{tb}{lst}{}
\floatname{listing}{Listing}
\title{Multi-Granular Multimodal Clue Fusion for Meme Understanding}

\author{
    Li Zheng\textsuperscript{\rm 1},
    Hao Fei\textsuperscript{\rm 2},
    Ting Dai\textsuperscript{\rm 1},
    Zuquan Peng\textsuperscript{\rm 1},
    Fei Li\textsuperscript{\rm 1,3}\thanks{Corresponding author.},
    Huisheng Ma\textsuperscript{\rm 4}, \\
    Chong Teng\textsuperscript{\rm 1},
    Donghong Ji\textsuperscript{\rm 1}
}
\affiliations{
    \textsuperscript{\rm 1}Key Laboratory of Aerospace Information Security and Trusted Computing, Ministry of Education,\\ School of Cyber Science and Engineering, Wuhan University, Wuhan, China\\
    \textsuperscript{\rm 2}National University of Singapore, Singapore, Singapore\\ 
    \textsuperscript{\rm 3}Laboratory for Advanced Computing and Intelligence Engineering, Wuxi, China \\
    \textsuperscript{\rm 4}North China Institute of Computing Technology, beijing, China

\{zhengli,daiting\_cs,pzq\_cse,lifei\_csnlp,tengchong,dhji\}@whu.edu.cn
\\
haofei37@nus.edu.sg, mhs@bupt.cn
}
\begin{document}
\maketitle
\begin{abstract}
With the continuous emergence of various social media platforms frequently used in daily life, the multimodal meme understanding (MMU) task has been garnering increasing attention. 
MMU aims to explore and comprehend the meanings of memes from various perspectives by performing tasks such as metaphor recognition, sentiment analysis, intention detection, and offensiveness detection. 
Despite making progress, limitations persist due to the loss of fine-grained metaphorical visual clue and the neglect of multimodal text-image weak correlation. 
To overcome these limitations, we propose a multi-granular multimodal clue fusion model (MGMCF) to advance MMU. 
Firstly, we design an object-level semantic mining module to extract object-level image feature clues, achieving fine-grained feature clue extraction and enhancing the model's ability to capture metaphorical details and semantics.
Secondly, we propose a brand-new global-local cross-modal interaction model to address the weak correlation between text and images. 
This model facilitates effective interaction between global multimodal contextual clues and local unimodal feature clues, strengthening their representations through a bidirectional cross-modal attention mechanism. 
Finally, we devise a dual-semantic guided training strategy to enhance the model's understanding and alignment of multimodal representations in the semantic space.
Experiments conducted on the widely-used MET-MEME bilingual dataset demonstrate significant improvements over state-of-the-art baselines. Specifically, there is an 8.14\% increase in precision for offensiveness detection task, and respective accuracy enhancements of 3.53\%, 3.89\%, and 3.52\% for metaphor recognition, sentiment analysis, and intention detection tasks.
These results, underpinned by in-depth analyses, underscore the effectiveness and potential of our approach for advancing MMU.

\end{abstract}
    
\section{Introduction}
Memes, as a popular form of online communication, express viewpoints, sentiments, and intentions in a concise and humorous manner. 
With the development of social networks, Multimodal Meme Understanding (MMU) \cite{wang2024they,DBLP:conf/sigir/XuLZNZL022}, as an emerging research area in Natural Language Processing (NLP), plays a crucial role in many downstream applications, such as question answering \cite{zheng2024reverse} and sentiment analysis \cite{zheng2023ecqed,zheng2023bi}.
The definition of the MMU task involves predicting understanding from four dimensions: metaphor, sentiment, intention, and offensiveness.
However, memes are nuanced, and accurately grasping the underlying meaning embedded within the combination of text and images poses a crucial challenge.

Several studies have made commendable efforts in MMU. 
\citet{kiela2020hateful,kirk2021memes} introduced multimodal hate meme datasets specifically designed for hate detection. However, these studies overlooked the crucial aspect of metaphorical features in memes. 
Therefore, \citet{DBLP:conf/sigir/XuLZNZL022} considered the richer metaphorical features in memes and constructed a baseline model and a bilingual dataset called MET-MEME for this purpose.
Furthermore, \citet{wang2024they} proposed a metaphor-aware multimodal multi-task framework on this dataset to capture the interactions between text and images. 
Despite achieving notable success, current researches in this field face two significant limitations: 1) \textbf{the loss of fine-grained metaphorical visual clues} and 2) \textbf{the neglect of multimodal text-image weak correlation}. These limitations hinder its further flourishing and widespread adoption.

 \begin{figure}[!t]
    \centering
    \includegraphics[scale=0.53]{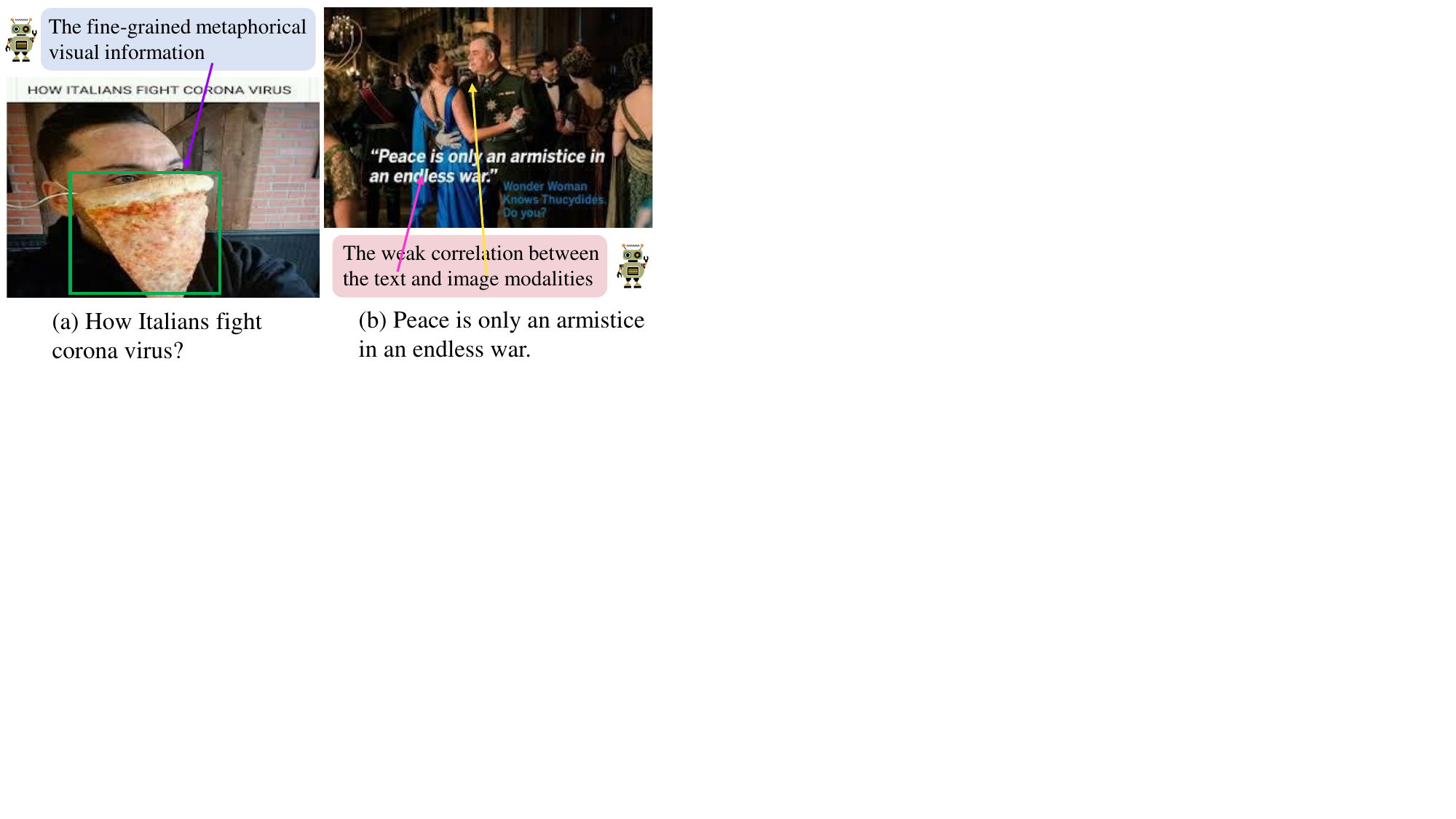}
    \caption{Examples of Metaphorical Memes.
    }
    \label{fig:ex}
    \vspace{-0.4cm}
\end{figure}

On the one hand, existing works~\cite{DBLP:conf/sp/QuHPBZZ23,ji2023identifying} exhibit a lack of emphasis on images and simply encode broad visual representations at the image-level, ignoring metaphorical clues at the fine-grained object-level of images.
This neglect leads to a critical absence of key visual metaphorical details, resulting in semantic ambiguity and omissions, ultimately failing to comprehensively capture the complexity and diversity of memes.
As shown in Figure \ref{fig:ex} (a), encoding visual features solely at the image-level falls short in capturing the crucial metaphorical clues of a pizza being used as a mask. 
Detecting the presence of metaphors and accurately predicting the conveyed sentiments, intentions, and potential offensiveness becomes exceedingly challenging in such cases.

On the other hand, existing methods~\cite{DBLP:conf/sigir/XuLZNZL022,wang2024they} primarily focus on directly integrating textual and visual information to comprehend memes, overlooking the issue of weak correlations between modalities and disregarding the intrinsic crucial metaphorical clues within each modality. 
This oversight lead to the loss of crucial details and clues, thereby limiting a comprehensive understanding of memes.
For instance, in the illustrated example in Figure \ref{fig:ex} (b), there is a weak correlation between the image and text, where the image conveys peace while the text reveals hatred towards war. 
Merely concatenating and fusing the image and text information directly could lead to a misinterpretation of the meme as peaceful.

In this paper, motivated by the aforementioned observations, we propose a \textit{\underline{\textbf{M}}ulti-\underline{\textbf{G}}ranular \underline{\textbf{M}}ultimodal \underline{\textbf{C}}lue \underline{\textbf{F}}usion model (MGMCF)} to improve multimodal meme understanding.
\textbf{First}, we design an object-level semantic mining module to extract fine-grained object-level feature clues from images.
We then integrate these object-level feature clues with the overall image-level feature clues to obtain a multi-granular representation.
This enables our model to better capture the metaphorical details and semantics of images, offering a more comprehensive visual understanding.
\textbf{Second}, considering the weak correlation between text and images, we not only focus on the interactions between different modalities but also emphasize the crucial metaphorical clues within each modality, integrating multi-granular clues to enhance the ultimate understanding of multimodal memes.
Therefore, we propose a novel global-local cross-modal interaction model to enable effective interaction between the global multimodal contextual clues and local unimodal feature clues. 
Specifically, the global multimodal context enhances the local unimodal features through a symmetric cross-modal attention mechanism. 
This interaction process is bidirectional, allowing the global context to extract useful clues from the local unimodal features and update itself. 
Through multi-level stacking, the global multimodal context and local unimodal features mutually enhance each other and gradually improve.
\textbf{Moreover}, to enhance semantic alignment, we devise a dual-semantic guided training strategy. 
By bringing related image-text pairs closer in the forward direction and pushing unrelated pairs apart in the reverse direction, we aim to foster a more robust understanding of complex multimodal semantic clues.

To verify the effectiveness of our model, we conduct experiments on the benchmark MET-MEME bilingual dataset \cite{DBLP:conf/sigir/XuLZNZL022}, which contains both English and Chinese memes. 
The results demonstrate that our model significantly outperforms all state-of-the-art (SoTA) baselines across all evaluation metrics in the four tasks. 
On the English meme dataset, the precision in the offensiveness detection task increased by 8.14\%, and the accuracy in metaphor recognition, sentiment analysis, and intention detection tasks improved by 3.53\%, 3.89\%, and 3.52\%, respectively. 
Additionally, extensive experiments validate the effectiveness of our fine-grained visual information enhancement and global-local interactions.

Our main contributions are summarized as follows:

\begin{itemize}

\item We analyze and summarize two intrinsic challenges in the MMU task and propose a multi-granular multimodal clue fusion model, the first to consider fine-grained visual clues and integrate unimodal feature clues to enhance MMU.

\item We design an object-level semantic mining module and a global-local cross-modal interaction model to facilitate effective interaction between global multimodal clues and local unimodal clues, achieving multi-granular understanding of meme metaphorical clues and semantics.

\item Our extensive experimental results on MET-MEME dataset demonstrate that our scheme achieves state-of-the-art performance on the MMU task.

\end{itemize}

\section{Related Work}

\subsection{Multimodal Meme Understanding}

Recently, multimodal meme understanding \cite{lin2024towards,DBLP:conf/ijcai/HeeCL23,DBLP:conf/sp/QuHPBZZ23,fang2024not,fang2023you,fang2024fewer,fang2025rethinking,zheng2024self} has attracted increasing attention. Unlike general multimodal learning tasks \cite{ji2021improving,ji2022knowing,li2022compositional,li2022fine,wu2023information,li2023revisiting,fei2024video,luo2024panosent}, meme understanding relies more heavily on contextual and metaphorical information. Existing research has mainly focused on hateful memes.
\citet{yang2023invariant} proposed a scalable invariant and specific modality representation learning framework based on graph neural networks for harmful meme detection. \citet{ji2023identifying} introduced a prompt-based method to identify harmful memes.
Beyond just focusing on hateful meme detection, \citet{DBLP:conf/sigir/XuLZNZL022}  introduced a fine-grained multimodal meme understanding dataset, which includes tasks such as sentiment analysis, intention detection, offensiveness detection, and metaphor recognition, to analyze memes at a finer granularity. \citet{wang2024they} created cross-modal and intra-modal attention mechanisms on this dataset to capture the interactions between text and images for multimodal meme understanding. 
However, existing works overlook the object-level fine-grained clues in images, potentially leading to the loss of crucial metaphorical visual details. Moreover, these methods do not address the issue of cross-modal weak correlations, neglecting the essential clues within unimodal, which result in semantic ambiguity and confusion, failing to fully capture the complexity and diversity of memes.

\begin{figure*}[!h]
    \centering
    \includegraphics[scale=0.4]{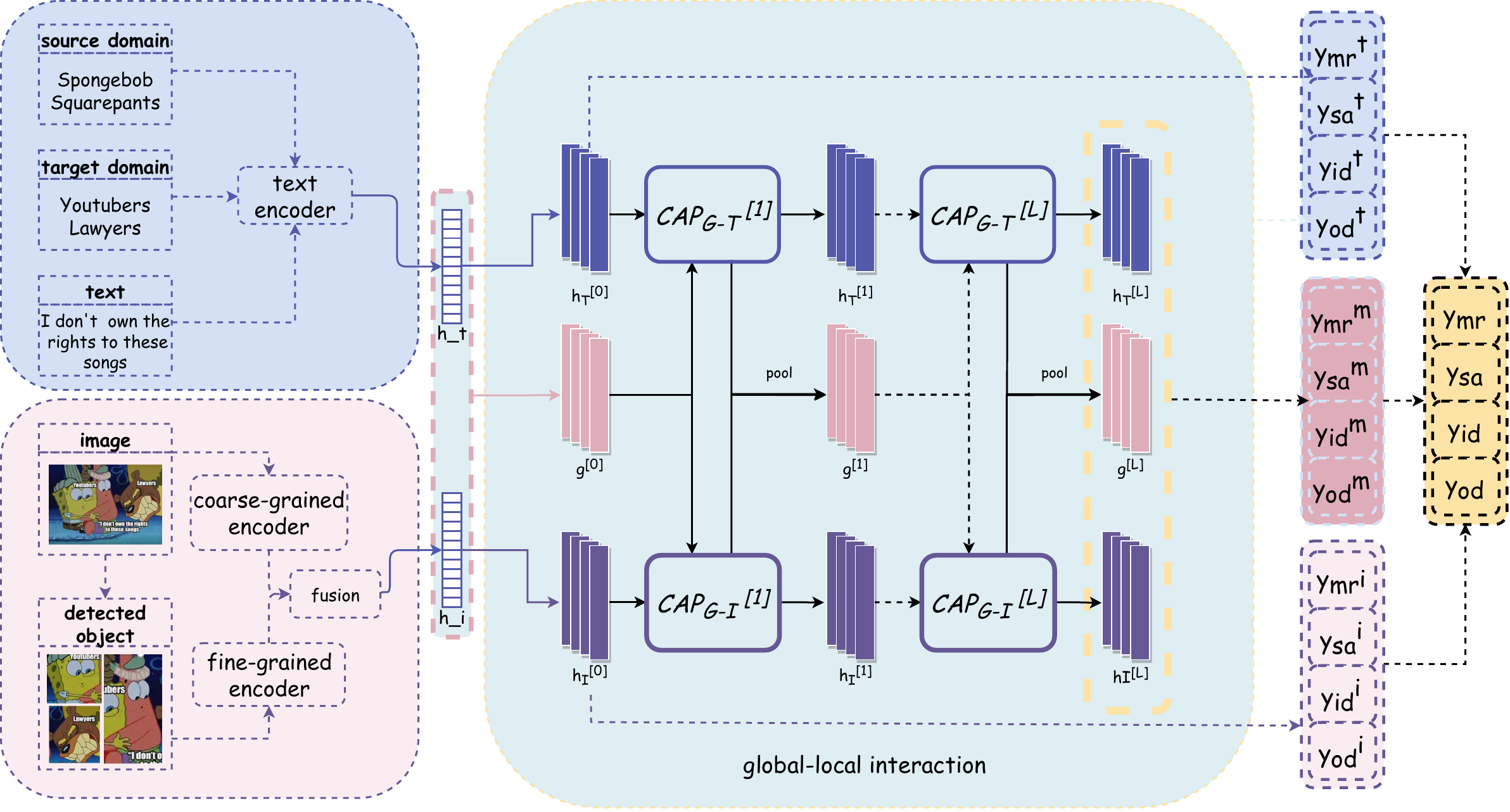}
    \caption{
    The overall architecture of our model. ``mr'' means metaphor recognition, ``sa'' means sentiment analysis, ``id'' means intention detection, ``od'' means offensiveness detection.}
    \label{fig:model}
    \vspace{-4mm}
\end{figure*}

\subsection{Metaphorical Information Processing}

In the field of NLP, there has been growing interest in developing models for metaphor detection \cite{DBLP:journals/ipm/HeYTYLW24,fang2024your,DBLP:conf/coling/ElzohbiZ24,DBLP:conf/acl/ZhangL23a}. Understanding the essence of memes relies critically on identifying the metaphorical information embedded within them. Existing researches \cite{DBLP:conf/coling/QiaoZM24,DBLP:conf/coling/ElzohbiZ24,DBLP:conf/acl/BadathalaKSB23} have primarily focused on unimodal metaphor detection.
\citet{DBLP:conf/acl/ZhangL23a} proposed a novel multi-task learning framework based on a metaphor recognition program, a set of linguistic rules.
\citet{DBLP:conf/acl/0001WLG23} performed metaphor detection by explicitly modeling the basic meanings of concepts.
\citet{DBLP:conf/emnlp/TianXM023} designed a domain contrastive learning strategy to capture the semantic inconsistencies. 
While these unimodal metaphor detection methods have achieved promising results, there has been relatively less exploration in the area of multimodal metaphor detection.
\citet{alnajjar2022ring} introduced a multimodal metaphor annotated corpus and designed a video-text content-based method for metaphor detection.
\citet{DBLP:journals/ipm/HeYTYLW24} developed a multi-interactive cross-modal residual network for multimodal metaphor recognition.

\section{Methodology}
\subsection{Task Definition}
This paper addresses the task of multimodal meme understanding, encompassing metaphor recognition,  sentiment analysis, intention detection, and offensiveness detection. 
Specifically, given an example consisting of an image $I$, its corresponding text $T$, a source domain $T_s$, and a target domain $T_a$, the objective of multimodal meme analysis is to predict the categories of metaphor $y_{mr}$, sentiment $y_{sa}$, intention $y_{id}$, and offensiveness $y_{od}$. 
As shown in Figure \ref{fig:model}, the source domain serves as the basis for the metaphor, while the target domain embodies the concept or idea metaphorically conveyed, typically in textual form.

\subsection{Feature Extraction}

\textbf{Text Encoder.}
In accordance with the approach described in \citet{wang2024they}, we utilize Multilingual BERT \cite{wang2019cross} to extract textual features from the corresponding text $x^t$, source domain $x^s$, and target domain $x^a$.
The encoding process can be formulated briefly as:
\begin{equation} 
\begin{split}
\{ \bm{h}_1^t,...,\bm{h}_n^t   \} = M\text{-}BERT( \{ \bm{x}_1^t,...,\bm{x}_n^t  \}  ) \\
\{ \bm{h}_1^s,...,\bm{h}_p^s   \} = M\text{-}BERT ( \{ \bm{x}_1^s,...,\bm{x}_p^s \} ) \\
 \{ \bm{h}_1^a,...,\bm{h}_q^a  \} = M\text{-}BERT ( \{ \bm{x}_1^a,...,\bm{x}_q^a \}  )
\end{split}
\end{equation}
where n, p, and q represent the word counts of the corresponding text, source domain, and target domain, respectively.

\noindent\textbf{Image Encoder.}

Multimodal meme images contain rich metaphorical details that are crucial clues for understanding memes. 
Therefore, when extracting visual features from memes, we cannot solely focus on image-level visual semantic clues as with other multimodal tasks. 
It is imperative to capture object-level fine-grained clue features that encompass these metaphorical details.
To achieve this goal, we devise a visual information enhancement strategy for extracting feature clues of different granularities.

For image I, following \cite{wang2024they}, we first employ a pretrained convolutional neural network classifier, VGG16 \cite{simonyan2014very}, to extract image-level features $\bm{h}^c = VGG16(I)$. 
Then, we transform the input image I into a series of embedded blocks to capture fine-grained image features.
By integrating object detection, attribute recognition, and positional information, we enrich the representation of image features and enhance enhance image metaphor comprehension. 
Specifically, we design an object-level semantic mining module \cite{anderson2018bottom} to identify and localize objects in an image.
For each visual region $I_i$ represented by a bounding box, we resize the region to a standard size of 224$\times$ 224 pixels. 
Following \citet{xu2020reasoning}, we reshape the resized region $I_i$ into a sequence $I_i = \{r_1, ..., r_m\}$, where each region is represented by a block.
This reshaping divides the region into a grid of blocks, with m being the total number of blocks.
Next, we flatten each block $r_j$ and project it into a $d^I$-dimensional vector. 
This projection is performed using a trainable linear projection matrix E, and the resulting embedded representation of block $r_j$ is denoted as $z_j = r_jE$.
To incorporate contextual information and retain positional information, we prepend a [class] token embedding at the beginning of the patch sequence. Position embeddings are also appended to the patch embeddings, indicating their relative positions within the sequence. The input representation of each visual region $I_i$ is expressed as:
\begin{equation} 
\bm{Z}_i = [\bm{z}_{[class]};\bm{z}_1,...,\bm{z}_m] + \bm{E}_{pos}
\end{equation}
where $Z_i$ represents the input matrix of image patches, and $E_{pos}$ denotes the position embedding matrix.
Subsequently, we feed the input matrix $Z_i$ into the VGG16 encoder to obtain the visual region $I_i$ representation $h_i^v = VGG16(Z_i)$.
Finally, the representation of the image I is defined as:
\begin{equation} 
\bm{h}_I = \{\bm{h}^c, \bm{h}_1^v,...,\bm{h}_m^v \}
\end{equation}

\subsection{Modal Fusion}

The text and image of multimodal meme have the problem of weak correlation, and directly fusing the text and image may result in incorrect meme understanding.
A good fusion solution should extract and integrate sufficient information from multimodal sequences while preserving the independence of each modality.
Therefore, we propose a novel global-local cross-modal interaction model that not only considers interactions between modalities but also emphasizes the importance of each modality itself to enhance multimodal fusion at multiple granularities.
Specifically, we devise an efficient mechanism called Cross-modal Attention Promotion (CAP) that leverages symmetric cross-modal attention to explore the inherent correlations between the two input feature sequences, promoting the exchange of beneficial information across the sequences.
CAP utilizes self-attention to model the temporal dependencies within each feature sequence, enabling the integration of more information. 
The mechanism takes sequences $\bm{h}^T$ and $\bm{h}^I$ as inputs and generates their mutually reinforcing information $\bm{h}_{T\to I}$ and $\bm{h}_{I\to T}$. 
The computation of $CAP_{T\leftrightarrow I}(\bm{h}^T, \bm{h}^I)$ is as follows:
\begin{equation} 
\begin{split}
\bm{h}_{T \to I}' = MCA(LN(\bm{h}_T),LN(\bm{h}_I)) + \bm{h}_T \\
\bm{h}_{T \to I}'' = MSA(LN(\bm{h}_{T \to I}')) + \bm{h}_{T \to I}' \\
\bm{h}_{T \to I} = FN(LN(\bm{h}_{T \to I}'')) + \bm{h}_{T \to I}''
\end{split}
\end{equation}
where LN denotes layer normalization and FN is the feedforward neural network.
$MSA(\cdot )$ refers to the output of the multi-head self-attention mechanism computation.
$MCA(\cdot, \cdot) $represents the result of the multiple cross-attention mechanism calculation.
Similarly, we can obtain $CAP_{I\leftrightarrow T}(\bm{h}_I, \bm{h}_T)$.

The traditional cross-attention interaction requires two updates during the modal interaction process to achieve modal enhancement, which is inefficient and introduces redundant features into the sequence. 
Based on the historical experience from large-scale pretraining, it has been observed that a single token can represent the entire sequence, further improving the efficiency of modal interaction. 
Motivated by this observation, we propose a global-local cross-modal interaction model with linear computational cost to enhance efficiency.
The discourse-level representation of each modality replaces the standard information and interacts with local unimodal features within a global multimodal context.
This means that the representation of each modality not only relies on local features but also takes into account the influence of global context. 
This global-local interaction model reduces the introduction of redundant features and improves modal interaction effectiveness while maintaining efficiency.

We establish the global multimodal context information denoted as $\bm{g}^i= concat(\bm{h}_T^i,\bm{h}_I^i)$ by concatenating the representations of each modality at each layer of global-local interaction, where i represents the number of layers of global local interaction.
By integrating the global context information and local modal information, and learning modal consistency and specificity, we ensure effective interaction and capture relevant information from both the global and local perspectives. The entire interaction process is as follows:
\begin{equation} 
\begin{split}
\bm{h}_T^{(i+1)},\bm{g}_{T \to G}^{(i)} = CAP_{T\leftrightarrow G}^{(i)}(\bm{h}_T^{(i)},\bm{g}^{(i)}) \\
\bm{h}_I^{(i+1)},\bm{g}_{I \to G}^{(i)} = CAP_{I\leftrightarrow G}^{(i)}(\bm{h}_I^{(i)},\bm{g}^{(i)})
\end{split}
\end{equation}
By stacking multiple layers, the global multimodal context and local unimodal features can mutually reinforce and progressively refine each other.
We hierarchically handle the entire learning process, with each layer capturing different features corresponding to the model's main stages.
The model initially learns shallow interaction features, gradually progressing to acquire higher-order semantic features in later stages. 
This hierarchical learning method successfully integrates information from various modalities by ingeniously designed aggregation blocks, providing the model with a more comprehensive and enriched representation of multimodal features.
Through the model's interactions, information from different modalities can be combined in a deeper and more effective manner, enabling the acquisition of more advanced feature representations in subsequent hierarchical learning. Subsequently, we aggregate the features from both unimodal and multimodal sources to facilitate subsequent task predictions.
\begin{equation} 
y_m = softmax(\bm{W}_mMSA([\bm{h}_T^{(L)}, \bm{h}_I^{(L)}, \bm{g}^{(L)}]) + \bm{b})
\end{equation}
where $y_m$ is the feature output distribution after multimodal fusion,
$W_m$ and $b$ are trainable parameters.

Our approach focuses not only on the interactions between modalities but also on the individual feature representations of each modality. 
We separately use the unimodal features obtained from text and image encoders to predict subsequent tasks.
This allows to capture the unique characteristics and information within each modality, thereby improving the accuracy and effectiveness of MMU.
\begin{equation} 
\begin{split}
\bm{y}_T = softmax(\bm{W}_t\bm{h}_T + \bm{b}) \\
\bm{y}_I = softmax(\bm{W}_i\bm{h}_I + \bm{b})
\end{split}
\end{equation}
Given the $y_M$, $y_T$, and $y_I$, we obtain the final prediction y:
\begin{equation} 
\bm{y} = \bm{y}_M + \bm{y}_T + \bm{y}_I
\end{equation}
where y can be considered as a comprehensive feature set encompassing multi-granular features, including text, image, and image-text modalities.

\begin{table*}[htp!]
\centering
\resizebox{\textwidth}{!}{
\renewcommand{\arraystretch}{1}
\begin{tabular}{lccccccccccccccc}
\hline
\multirow{2}[0]{*}{\textbf{Method}} & \multicolumn{3}{c}{\textbf{English}} & &\multicolumn{3}{c}{\textbf{Chinese}} & & \multicolumn{3}{c}{\textbf{English}} & &\multicolumn{3}{c}{\textbf{Chinese}} 
\\
\cline{2-4} \cline{6-8} \cline{10-12} \cline{14-16}

 & \textbf{Acc} & \textbf{Pre} & \textbf{Rec} & & \textbf{Acc} & \textbf{Pre} & \textbf{Rec} & &  \textbf{Acc} & \textbf{Pre} & \textbf{Rec} & & \textbf{Acc} & \textbf{Pre} & \textbf{Rec} 
 \\
\hline
 & \multicolumn{7}{c}{\textbf{Sentiment Analysis}} & &  \multicolumn{7}{c}{\textbf{Intention Detection}} \\
 \cline{2-8}\cline{10-16}
 
VGG16 & 20.57 & 20.84 & 24.22 & & 29.94 & 26.04 & 29.20 & &  37.19 & 38.71 & 38.15  & & 47.48 & 49.21 & 47.81 \\
DenseNet-161 & 21.88 & 21.71 & 25.65  & & 29.45 & 27.50 & 29.36 & &  38.10 & 39.31 & 37.89  & & 47.23 & 39.24 & 47.06 \\
ResNet-50 & 21.74 & 18.63 & 21.35 & & 29.36 & 27.50 & 29.28 & &  39.19 & 37.12 & 40.10  & & 47.15 & 39.23 & 47.06 \\
Multi-BERT\_EfficientNet & 28.52 & 24.52 & 29.04 & & 33.50 & 35.29 & 33.42 & &  43.10 & 41.54 & 42.19  & & 51.03 & 43.06 & 51.03 \\
Multi-BERT\_ViT & 24.43 & 23.41 & 23.96 & & 33.25 & 27.33 & 32.84 & &  41.28 & 40.13 & 40.62  & & 50.62 & 41.32 & 50.62 \\
Multi-BERT\_PiT & 25.00 & 27.82 & 28.12 & & 33.66 & 33.58 & 33.09 & &  42.23 & 41.09 & 41.02  & & 50.21 & 50.00 & 50.04 \\

MET\_add & 24.65 & 24.52 & 25.26&  &32.50 & 32.62 & 33.50& & 40.32 & 40.39 & 41.28&  &52.93 & 52.68 &54.01\\
MET\_cat & 27.68 & 28.41 & 29.82 &  &33.42 & 34.33 & 33.91 & & 38.56 & 39.19 & 39.84&  &51.58 & 51.48 & 52.85 \\
M3F\_add & \uline{30.47} & 33.45 & 30.34 &  &\uline{39.95} & \uline{41.80} & \uline{39.87}& & \uline{44.40} & 41.89 & \uline{44.32} &  &\uline{55.25}& 54.57 & \uline{55.00} \\
M3F\_cat & 29.82 & \uline{34.18} & \uline{30.73}& &37.22 & 39.55 &37.97 & & 44.10 & \uline{44.56} & 43.53& &53.52 & \uline{54.72} & 54.52\\
Ours & \textbf{34.36}& \textbf{37.77}& \textbf{34.38}&& \textbf{42.11}& \textbf{47.59}& \textbf{42.02} && \textbf{47.92}& \textbf{47.53}& \textbf{47.06}&& \textbf{58.56}& \textbf{57.99}& \textbf{58.32}  \\
 \specialrule{0em}{-2pt}{-1pt}   & \scriptsize{(+3.89\%})  & \scriptsize{(+3.52\%})  &  \scriptsize{(+3.65\%}) && \scriptsize{(+2.16\%})  & \scriptsize{(+5.79\%})  &  \scriptsize{(+2.15\%}) && \scriptsize{(+3.52\%})  & \scriptsize{(+2.97\%})  &  \scriptsize{(+2.74\%}) && \scriptsize{(+3.31\%})  & \scriptsize{(+3.27\%})  &  \scriptsize{(+3.32\%})  \\
\hline

& \multicolumn{7}{c}{\textbf{Offensiveness Detection}} & &\multicolumn{7}{c}{\textbf{Metaphor Recognition}} \\
  \cline{2-8}\cline{10-16}

VGG16 & 67.10 & 63.42 & 72.53 && 70.07 & 64.11 & 72.07 && 78.39 & 79.73 & 79.95 && 67.00 & 67.24 & 67.82 \\
DenseNet-161 & 69.66 & 62.07 & 69.98 && 71.43 & 70.82 & 75.43 & &80.08 & 80.23 & 80.47 && 67.16 & 67.91 & 67.99 \\
ResNet-50 & 69.21 & 64.62 & 72.57 && 73.24 & 69.62 & 75.74 & &80.34 & 81.22 & 80.86 && 67.74 & 67.63 & 67.66 \\
Multi-BERT\_EfficientNet & 73.78 & 67.98 & 74.56 && 78.15 & 72.11 & 79.98 && 82.46 & 84.39 & 83.11 && 74.19 & 71.26 & 74.28 \\
Multi-BERT\_ViT & 71.22 & 62.96 & 72.66 && 76.92 & 67.31 & 78.74 & &81.90 & 82.01 & 82.46 && 73.28 & 72.55 & 73.70 \\
Multi-BERT\_PiT & 72.79 & 66.69 & 74.26 && 77.17 & 70.16 & 79.05 && 82.07 & 83.05 & 82.98 && 75.10 & 73.15 & 74.28 \\

MET\_add & 68.39 & 66.21 & 72.14 &  &76.01 & 74.76 & 78.16& & 81.33 & 81.49 & 82.29 & &74.04 &\uline{74.51} &74.96 \\
MET\_cat & 67.25 & 66.15 & 74.48&  &73.19 & 71.59 & 79.49& & 82.39 &82.69 & 83.33 &  &72.90 & 72.80 & 75.67 \\
M3F\_add & \uline{76.17} & 69.45 & \uline{76.19}&  &\uline{80.81} & 76.00 & \uline{80.73}& & \uline{83.98} & 85.86 &84.38&  &\uline{77.01}&72.94 &\uline{82.68}\\
M3F\_cat & 74.09 & \uline{69.59} &76.15& &80.07 & \uline{76.20} & 80.62& & 83.20 & \uline{85.97} & \uline{85.81} & &76.18 & 73.02 & 80.00 \\

Ours & \textbf{78.11}& \textbf{77.73}& \textbf{78.32} && \textbf{82.15}& \textbf{81.80}& \textbf{82.02} && \textbf{87.51}& \textbf{88.58} & \textbf{88.89}& & \textbf{78.39}& \textbf{78.16} & \textbf{83.92}\\
\specialrule{0em}{-2pt}{-1pt}   & \scriptsize{(+1.94\%})  & \scriptsize{(+8.14\%})  &  \scriptsize{(+2.13\%})  & & \scriptsize{(+1.34\%})  & \scriptsize{(+5.6\%})  &  \scriptsize{(+1.29\%})& & \scriptsize{(+3.53\%})  & \scriptsize{(+2.61\%})  &  \scriptsize{(+3.08\%})  & & \scriptsize{(+1.38\%})  & \scriptsize{(+3.65\%})  &  \scriptsize{(+1.24\%})\\
\hline
\end{tabular}}
\caption{Experimental results on the MET-MEME dataset of four tasks.}
\label{tab:main result}
\vspace{-4mm}
\end{table*}

\subsection{Training}
For each task, we train the model using the standard gradient descent algorithm to minimize the cross-entropy loss.
\begin{equation} 
\min_{\theta } \mathcal{L}_{*} = -\sum_{i = 1}^{N} y_{*}^ilog\hat{y}_{*}^i + \lambda_{*} \left \| \theta_{*}  \right \|^2 
\end{equation}
where $*$ stands for the representation of different tasks. 
N is the training data size. $y^i$ and $\hat{y}^i$ respectively represent the ground-truth and estimated label distribution of instance i. $\theta$ denotes all trainable parameters of the model, $\lambda$ represents the coefficient of L2-regularization.

\noindent\textbf{Dual-semantic Guided Loss.}
In addition to task-specific losses, we devise a dual-semantic guided loss to effectively leverage cross-modal information, enhancing the model's comprehension and alignment of multimodal representations in the semantic space. 
The context-aware multimodal representations ($\bm{h}{T_i}$ and $\bm{h}{I_I}$) contain contextually relevant information associated with specific memes. 
By bringing related image-text pairs closer in the forward direction and pushing unrelated pairs apart in the reverse direction, these representations are aligned in the same semantic space, effectively utilizing cross-modal information.
Specifically, we contrast the multimodal representation of a specific meme sample (i.e., $\bm{h}_{T_i}$), with another multimodal representation ($\bm{h}_{I_i}$) from within the same batch of sampled memes. 
By comparing the similarities and differences between these representations, the model learns how to better differentiate and capture the semantic information among different meme samples. 
\begin{equation} 
\mathcal{L}_{dg} = -log\frac{exp(sim(\bm{h}_{T_i},\bm{h}_{I_i})/\tau )}{ {\textstyle \sum_{k=1[k\ne i]}^{2N}} exp(sim(\bm{h}_{T_k},\bm{h}_{I_k})/\tau ))} 
\end{equation}
where sim is the cosine-similarity, N is the batch size, and $\tau$ is the temperature to scale the logits.

By minimizing task-specific losses for each individual task and incorporating a contrastive loss, our overall loss function is defined as follows:
\begin{equation} 
\mathcal{L} = \mathcal{L}_{mr} + \mathcal{L}_{sa} + \mathcal{L}_{id} + \mathcal{L}_{od} + \mathcal{L}_{dg}
\end{equation}

\section{Experiments}
\subsection{Experimental Setting}
\noindent\textbf{Datasets.} We assess the efficacy of our model on the widely used MET-MEME bilingual dataset \cite{DBLP:conf/sigir/XuLZNZL022}, which consists of both English and Chinese memes. 
The English meme dataset is sourced from MEMOTION and Google search, comprising 4,000 text-image pairs, comprising 4,000 text-image pairs.
The Chinese meme dataset consists of 6,045 text-image pairs, covering six different categories, including animals, scenery, animations, films, dolls, and humans.

\noindent\textbf{Evaluation Metrics.}
In terms of evaluation metrics, we align with \citet{wang2024they} and use three metrics, namely accuracy (Acc), weighted precision (Pre), and recall (Rec), to assess the performance.
All our scores are the average over 5 runnings with random seeds.

\subsection{Baseline Systems}

To validate the effectiveness of our model, we compare it against the following state-of-the-art baselines.
(1) Unimodal models that solely utilize image modality information, such as VGG16 \cite{simonyan2014very}, DenseNet-161 \cite{huang2017densely}, and ResNet-50 \cite{he2016deep}.
(2) Multimodal models that incorporate both text and image modalities. These include Multi-BERT-EfficientNet \cite{tan2019efficientnet}, Multi-BERT-ViT \cite{dosovitskiy2020image}, Multi-BERT-PiT \cite{heo2021rethinking}, MET \cite{DBLP:conf/sigir/XuLZNZL022} employ a straightforward concatenation or element-wise addition to merge feature vectors for meme understanding, and M3F \cite{wang2024they} devise attention mechanisms to capture the interaction between text and images.

\subsection{Main Results}

We conduct a comprehensive comparison of our MGMCF with SoTA unimodal and multimodal models on the MET-MEME dataset.
Table \ref{tab:main result} presents the results for four tasks: sentiment analysis (SA), intention detection (ID), offensiveness detection (OD), and metaphor recognition (MR). 
The results highlight that our approach outperforms SoTA baselines across all evaluation metrics for the four tasks, revealing several key findings.
Firstly, compared to unimodal models, multimodal models demonstrate superior performance by leveraging additional visual-textual features. 
However, it is crucial to thoroughly exploit visual information and deeply fuse multimodal features. 
Otherwise, they only yield marginal improvements over unimodal models or even underperform them (e.g., in the OD task, the MET method performs worse than DenseNet-161 and ResNet-50).
By creating cross-modal attention,  M3F achieves the current SoTA results. 
Notably, our model significantly surpasses the SoTA techniques. 
On the English meme dataset, our precision improves by 8.14\% in the OD task, and accuracy improves by 3.53\%, 3.89\%, and 3.52\% in MR, SA, and ID, respectively. 
On the Chinese dataset, our precision improves by 3.65\%, 5.79\%, 3.27\%, and 5.6\% in MR, SA, ID, and OD, respectively.
Furthermore, our approach exhibits significant enhancements compared to MET. 
On the English MEME dataset, improvements range from 6.18\% to 13.25\%, and on the Chinese MEME dataset, enhancements range from 3.86\% to 14.97\%.
These findings indicate that by focusing on object-level fine-grained image details and the intrinsic unimodal clues, and integrating multi-granular clues, we achieve significant performance improvements in the MMU.

\begin{table}[t!]
\centering
\renewcommand{\arraystretch}{1}
\resizebox{0.48\textwidth}{!}{
\begin{tabular}{lccccccc}
\hline
\multirow{2}[0]{*}{\textbf{Method}}
& \multicolumn{3}{c}{\textbf{Sentiment Analysis}}& & \multicolumn{3}{c}{\textbf{Intention Detection}}\\
\cline{2-4} \cline{6-8} 
& \textbf{Acc} & \textbf{Pre} & \textbf{Rec} && \textbf{Acc} & \textbf{Pre} & \textbf{Rec}\\
\hline
Ours & 34.36 & 37.77 & 34.38 & & 47.92 & 47.53 & 47.06 \\
w/o OM & 32.47  & 36.16 & 32.39 && 46.27  & 46.17 & 45.89 \\
w/o UP & 32.85  & 36.53& 32.76 && 46.73  & 46.59  & 46.11  \\
w/o GL & 31.76  & 35.49 & 31.83  && 45.82& 45.41  & 45.47 \\
w/o DG & 33.59  & 36.94  & 33.52  && 47.05  & 46.94  & 46.63 \\
\hline
& \multicolumn{3}{c}{\textbf{Offensiveness Detection}} & &\multicolumn{3}{c}{\textbf{Metaphor Recognition}} \\
\cline{2-4}\cline{6-8}
& \textbf{Acc} & \textbf{Pre} & \textbf{Rec} && \textbf{Acc} & \textbf{Pre} & \textbf{Rec}\\
\hline
Ours & 78.11 & 77.73 & 78.32 & & 87.51 & 88.58 & 88.89 \\
w/o OM & 77.28  & 73.41 & 77.22 & & 85.42 & 87.45  & 87.46 \\
w/o UP & 77.53  & 74.66 & 77.68  && 85.81  & 87.64  & 87.74 \\
w/o GL & 76.89  & 72.57 & 76.94  && 84.98  & 86.77 & 86.79 \\
w/o DG & 77.74  & 75.39  & 77.93 && 86.43  & 87.93  & 88.02  \\
\hline
\end{tabular}
}
\caption{Ablation results on the MET-MEME English dataset of four tasks. ``OM'' means object-level semantic mining, ``UP'' means
unimodal prediction, ``GL'' means global-local interaction, ``DG'' means dual-semantic guided strategy.
}
\label{tab:ablation}
\vspace{-4mm}
\end{table}

\begin{figure}
    \centering
    \includegraphics[width=0.9\linewidth]{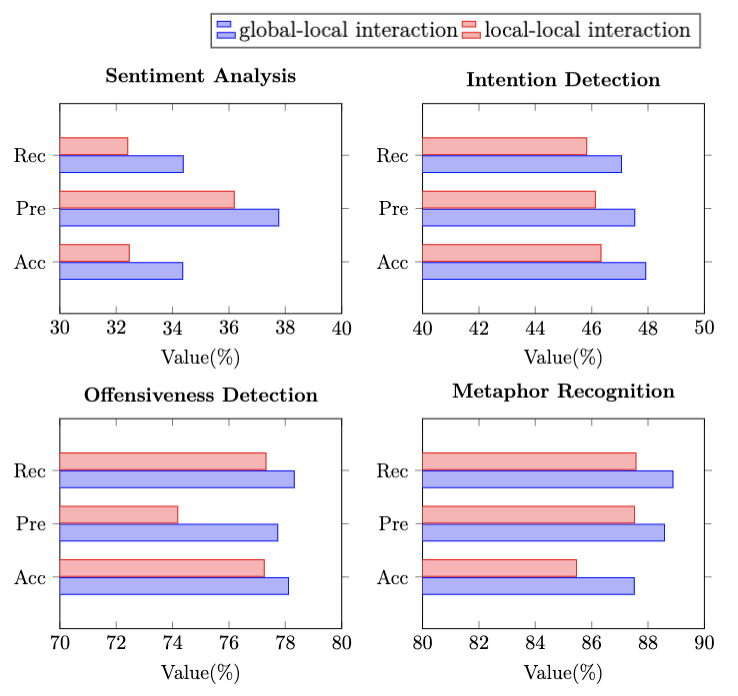}
    \caption{Comparative results between global-local and local-local interaction on the MET-MEME English dataset.}
    \label{fig:q1}
    \vspace{-4mm}
\end{figure}

\subsection{Ablation Study}
We perform ablation experiments to evaluate the contribution of each component in our model. 
As depicted in Table \ref{tab:ablation}, no variant matches the full model’s performance, highlighting the indispensability of each component. 
Specifically, when the global-local interaction is not utilized, three evaluation metric scores for all four tasks suffer the most significant decline. 
In particular, the precision score for the OD task drops by 5.16\%, and for the SA task drops by 2.28\%. 
This indicates that the global-local interaction successfully integrates information from different modalities, providing a more comprehensive multimodal feature representation.
To validate the necessity of object-level semantic mining module, we remove this module, and the decline in results signifies its indispensable impact on MMU. 
This finding suggests that mining fine-grained information from images can offer more detailed insights into image content. 
Furthermore, removing unimodal predictions leads to a performance drop, indicating that unimodal predictions contribute to the final predictions and effectively address the issue of weak correlation between image and text. 
Additionally, performance declines when dual-semantic guided strategy is removed, demonstrating its crucial role in enhancing semantic alignment and reducing modalities' inconsistencies.

\subsection{Deep Analyses on The Proposed Methods}
To further investigate the effectiveness of our method, we conduct in-depth analyses to answer the following questions, aiming to mine the intuition and analyze implicit phenomena.

\noindent\textbf{Q1: What are the advantages of the global-local
interaction?}
To further validate the effectiveness of our proposed global-local interaction model, we conduct a comparative analysis between our global-local interaction method and the local-local interaction method. 
The local-local interaction method refers to performing pairwise local interactions within each modality.
As shown in Figure \ref{fig:q1}, the results consistently demonstrate the superiority of the global-local interaction across all evaluation metrics for the four tasks. 
This finding indicates that by incorporating a higher-level global context that encompasses the entire modality, the global-local interaction method achieves more comprehensive and effective interactions between modalities. 
In contrast, the local-local interaction method solely focuses on intra-modality local feature interactions and fails to fully leverage the holistic information across modalities.
By leveraging the global context, we capture a broader range of semantic information, enabling a deeper understanding of the interdependence between modalities and effectively addressing inconsistencies and differences between modalities, thereby enhancing the performance of the MMU task.

\begin{figure}[!t]
    \centering
    \includegraphics[scale=0.3]{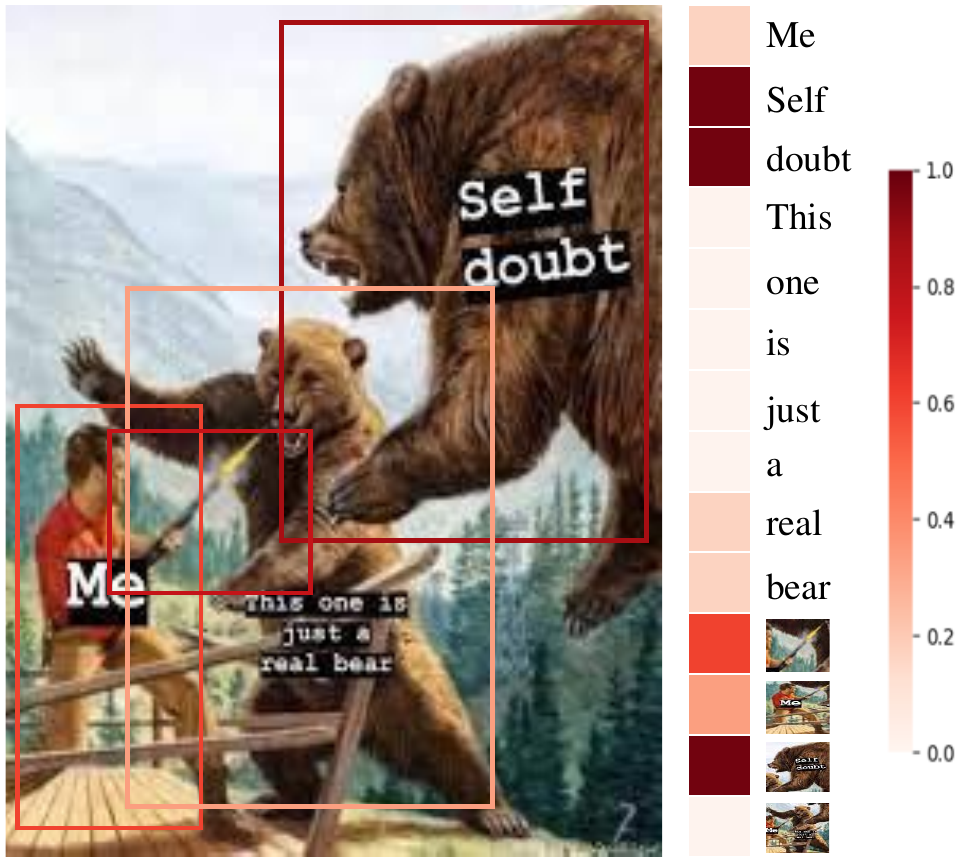}
    \caption{Visualization of a typical example.
    }
    \label{fig:visual}
    \vspace{-4mm}
\end{figure}

\begin{figure}
    \centering
    \includegraphics[width=1.0\linewidth]{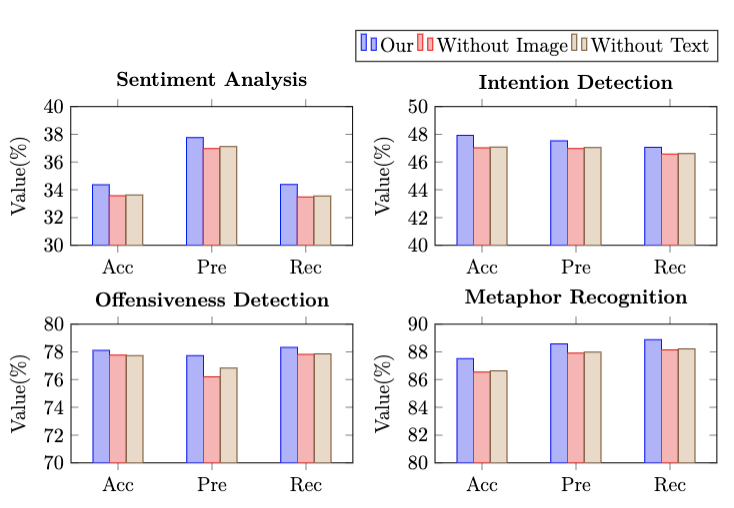}
    \caption{Influence of unimodal prediction on the MET-MEME English dataset.}
    \label{fig:single}
    \vspace{-4mm}
\end{figure}

\noindent\textbf{Q2: How can fine-grained visual features help improve model performance? }
We conduct a visualization in Figure \ref{fig:visual} to better illustrate the outstanding performance of the fine-grained visual enhancement module. 
By visualizing the attention distribution of the model, we observe that the module exhibits higher attention towards specific object parts in the image when the fine-grained visual enhancement module is utilized. Focusing on specific objects rather than the entire image allows for more accurate capture of crucial visual details. This confirms the effectiveness of the fine-grained visual enhancement module in improving the model's attention and understanding of key visual information.

\noindent\textbf{Q3: What are the advantages of unimodal prediction in multimodal meme understanding? }
In order to validate the effectiveness of incorporating unimodal prediction in multimodal meme understanding, we conduct a comparative analysis of the performance of combining unimodal prediction with separately removing text modal prediction and image modal prediction.
As illustrated in Figure \ref{fig:single}, the results consistently show that the combined unimodal prediction outperform the scenarios where any unimodal prediction is removed across all evaluation metrics in the four tasks. 
This indicates that by focusing on the crucial information within each modality, we can achieve a more comprehensive and accurate understanding of the visual and textual elements within memes.
Furthermore, we observe that removing the image modality prediction has a larger impact on the performance compared to removing the text modality prediction.
This finding emphasizes the importance of enhancing visual information. 
By capturing fine-grained visual details, the model can better leverage the rich visual cues and context present in memes. 
These findings highlight the value of unimodal prediction and underscore the significance of considering the specific characteristics of each modality in MMU.

\begin{figure}
    \centering
    \includegraphics[width=0.8\linewidth]{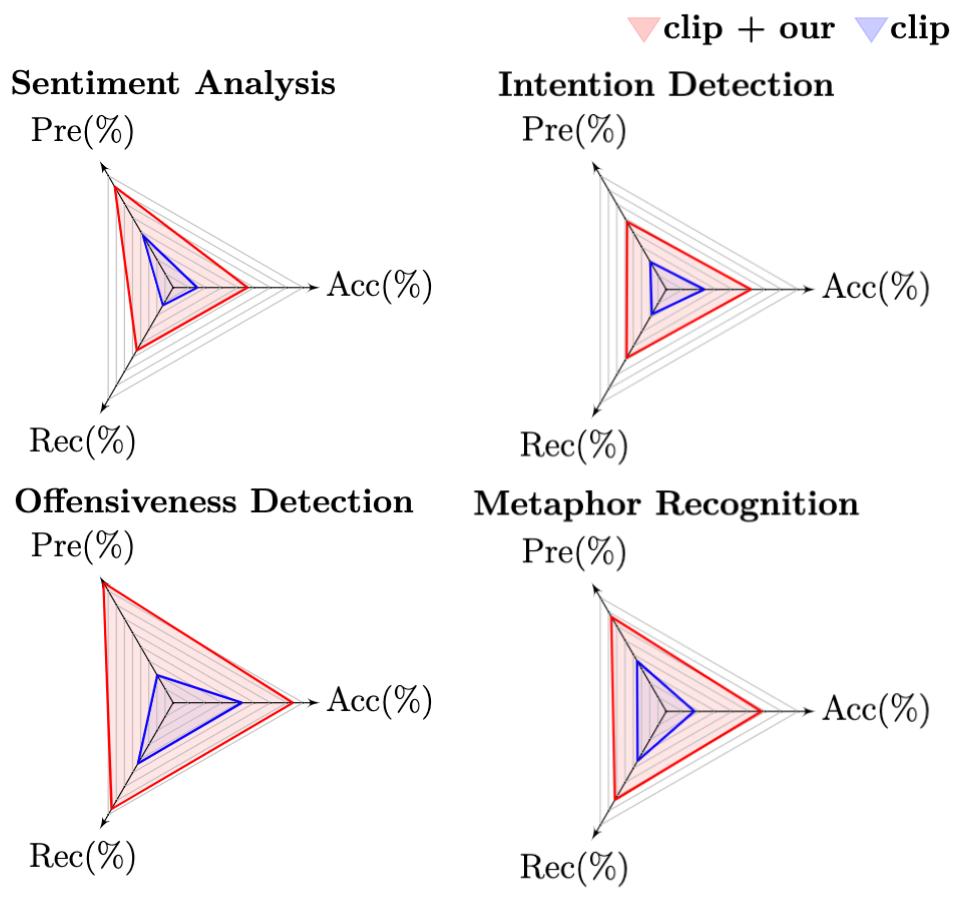}
    \caption{Influence of CLIP in MMU on the MET-MEME English dataset.}
    \label{fig:clip}
    \vspace{-6mm}
\end{figure}

\noindent\textbf{Q4: What impact does a large language model have on MMU?}
Considering the extensive usage of large language models \cite{wu2024towards,wu24next,fei2024enhancing,fei2024vitron}, we investigate their influence on MMU. 
Figure \ref{fig:clip} displays the results achieved by employing the standalone CLIP model and by integrating the CLIP model with our proposed method. 
Notably, employing the CLIP model alone yields impressive performance, attesting to its adeptness in comprehending multimodal memes. 
Moreover, the integration of the CLIP model with our method results in additional performance improvements, underscoring the efficacy of our approach.

\section{Conclusion}
In this paper, we explore two major challenges in the MMU task: the loss of fine-grained metaphorical visual clues and the neglect of weak correlation between multimodal text and images, proposing a solution named MGMCF. 
MGMCF enhances the complexity and diversity of images by capturing object-level fine-grained visual clues, and resolves the weak correlation between text and images through a novel global-local cross-modal interaction for multi-granular clue fusion. Extensive experiments on the MET-MEME bilingual dataset demonstrate the effectiveness of all our proposed innovative methods and hypotheses, achieving SoTA performance.

\section{Acknowledgments}
This work is supported by the National Key Research and Development Program of China (No. 2022YFB3103602), the National Natural Science Foundation of China (No. 62176187, No. 62202210). 
This work is also supported by the fund of Laboratory for Advanced Computing and Intelligence Engineering.

\bibliography{aaai25}

\end{document}